\documentclass[letterpaper, 10 pt, conference]{ieeeconf}  

\IEEEoverridecommandlockouts                              
\usepackage{times}
\usepackage{helvet}
\usepackage{courier}
\usepackage{latexsym}
\usepackage[english]{babel}

\usepackage{verbatim}
\usepackage{amsmath}
\usepackage{amsfonts}
\usepackage{graphicx}
\usepackage{relsize}
\usepackage{amssymb}  
\usepackage{algorithm}
\usepackage{algpseudocode}
\usepackage{listings}
\usepackage{longtable}
\usepackage{textcomp}
\usepackage{url}
\usepackage{alltt}
\usepackage{caption}
\usepackage{subcaption}
\usepackage{wrapfig}
\usepackage{multicol}
\usepackage{color}
\usepackage{cite}
\usepackage{epstopdf}

\title{\LARGE \bf
User Experience of the CoSTAR System \\ for Instruction of Collaborative Robots
}

\author{Chris Paxton,$^{1}$ Felix Jonathan,$^{1}$ Andrew Hundt,$^{1}$ Bilge Mutlu,$^{2}$ and Gregory D. Hager$^{1}$
\thanks{*This work was supported by NSF grant number 1637949.}
\thanks{$^{1}$Department of Computer Science, Johns Hopkins University, 3400 N Charles Street, Baltimore, MD 21218, USA
        {\tt\small cpaxton,fjonath1,ahundt1,hager@jhu.edu}}%
\thanks{$^{2}$Department of Computer Sciences, University of Wisconsin--Madison, 1210 W Dayton Street, Madison, WI 53706, USA 
        {\tt\small bilge@cs.wisc.edu}}%
}

\begin{document}

\maketitle
\thispagestyle{empty}
\pagestyle{empty}

\begin{abstract}

How can we enable novice users to create effective task plans for collaborative robots?
Must there be a tradeoff between generalizability and ease of use?
To answer these questions, we conducted a user study with the CoSTAR system, which integrates perception and reasoning into a Behavior Tree-based task plan editor. 
In our study, we ask novice users to perform simple pick-and-place assembly tasks under varying perception and planning capabilities.
Our study shows that users found Behavior Trees to be an effective way of specifying task plans. 
Furthermore, users were also able to more quickly, effectively, and generally author task plans with the addition of CoSTAR's planning, perception, and reasoning capabilities.
Despite these improvements, concepts associated with these capabilities were rated by users as less usable, and our results suggest a direction for further refinement.

\end{abstract}

\section{INTRODUCTION}

What will enable ordinary people to teach robots to solve difficult problems?
As robots become more important in industry and broader society, we must be able to answer this question. To this end, we are developing the open-source CoSTAR system\footnote{\url{https://github.com/cpaxton/costar_stack}} for authoring complex robot task plans~\cite{guerin2015costar,paxton2017costar}. CoSTAR integrates perception and reasoning capabilities into a cohesive platform that allows end-users to author robot programs.

Building a powerful interface that allows a non-expert user to create robot programs is a widely-pursued-goal in both industry and academia~\cite{nguyen2013ros,mateo2014hammer,guerin2015costar,steinmetz2016skill,dianov2016extracting}. Most collaborative robots, including the Universal Robots UR5 or the Rethink Robotics Sawyer, are packaged with simple tools for editing and designing task plans. Programs such as ROS Commander~\cite{nguyen2013ros} and RAFCON~\cite{steinmetz2016skill} allow users to combine finite state machines. By contrast, approaches based on symbols and ontologies such as~\cite{dianov2016extracting} aim to do more work for the user, but require a specially constructed setup for a particular problem made by experts before users can use it to solve their problem. In industry, Rethink Robotics revealed a new Behavior Tree-based user interface for programming their collaborative robots~\cite{intera5}, an indicator that this is a problem area in need of real solutions. 

\begin{figure}[bt]
\centering
\includegraphics[width=\columnwidth]{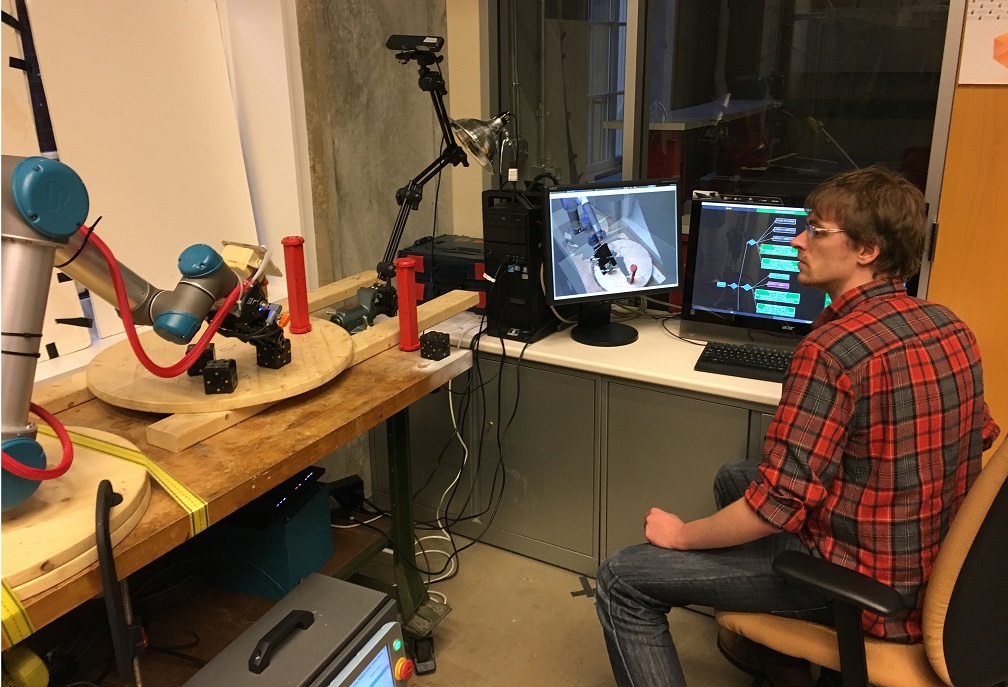}
\caption{The CoSTAR system performing a simple manipulation task. The current system includes a visualization of the robot's workspace, a Behavior Tree-based editor, a Primesense RGBD sensor, and a Universal Robots UR5 with a mounted Robotiq S-model gripper.}
\label{fig:intro}
\end{figure}

In our previous work~\cite{paxton2017costar}, we describe the three goals of a framework for authoring robot task plans: (1) it must be capable of solving a wide variety of problems; (2) it must be easily adaptable to new contexts; and (3) it must be robust to environmental variation.
In this work, we address the challenge of building a \textit{usable} system by this definition.
To accomplish this, our ideal system must be able to incorporate both perception and user knowledge, and it must do so in an intelligent way. Humans can solve problems that many robots cannot solve; conversely it is important that humans be able to develop an accurate mental model of what a robot is capable of to work with them accurately and efficiently~\cite{sharp2007interaction}.

We presented the CoSTAR system as a Behavior Tree-based editor~\cite{guerin2015costar,paxton2017costar}. Our initial version of the system had very limited planning or reasoning capabilities, and relied heavily on having a skilled roboticist for a user who could create elaborate task plans to solve any problems it faced. We then extended this system to be more reliable, capable and cross-platform, integrating perception and simple reasoning~\cite{paxton2017costar}.
The CoSTAR system's solution to these problems is to allow humans to solve problems through a combination of \textit{Operations} of varying degrees of complexity. At the lowest level, we have atomic operations that open a gripper or that servo an arm to a specific position.
Increased complexity allows users to generate motions to specific positions, to perform complex operations with known effects, and to query the system about the state of the world. 

While our previous work described the system as a whole~\cite{paxton2017costar}, 
the extent to which this system can support end users in instructing collaborative robots complex tasks was unknown.
%
In this paper, we focus on an evaluation of its effectiveness in the hands of novice users and how this information can inform the development of such systems going forward.
For our preliminary study, we examine a simple structure assembly task that has been explored in depth in the past~\cite{paxton2016want,paxton2017costar}.
We separate the CoSTAR systems into four sets of abilities that correspond to, (1) \textit{Baseline}: a basic system similar to many commercially available systems with only the ability to servo to pre-programmed waypoints, (2) \textit{Planning}: a system similar to the above but that incorporates motion planning to avoid obstacles, (3) \textit{Perception}: a system that uses perception but no planning, and (4) \textit{SmartMove}: a system integrating perception and planning through a set of abstract queries called SmartMoves.

To summarize, the contributions of this work are:
(1) an expansion of the open-source CoSTAR system to allow novices to author task plans for collaborative robots, (2) a preliminary study examining which characteristics of this system most enhance effectiveness and user experience.
This study gives us new insight into the design of future user interfaces for instructing collaborative robots.

\section{BACKGROUND}

There is strong interest in developing systems and user interfaces that allow non-expert users to program robots. \cite{goodrich2007hrisurvey,steinfeld2006metrics,albert2009beyond} each provide an overview of various areas relevant to Human Robot Interaction, which also has roots in Interaction Design~\cite{sharp2007interaction}.
Recent approaches relevant to the robotics domain include both new user interfaces, e.g. in~\cite{nguyen2013ros,mateo2014hammer,guerin2015costar,steinmetz2016skill}, learning from demonstration~\cite{ahmadzadeh2015learning}, or systems that make use of onotologies and bases like Tell Me Dave~\cite{misra2014tell} or
RoboSherlock~\cite{beetz2015robosherlock,worch2016perception}.

Our proposed user interface is based on Behavior Trees, which have previously been used on humanoid and surgical robots, among other applications~\cite{bagnell2012integrated,marzinotto2014towards,hu2015semi}.
The most relevant prior work describes the previous version of the CoSTAR system~\cite{paxton2017costar}, which consists of a cross-platform framework and an architecture for incorporating abstract knowledge and perception into a user-friendly system. A Behavior Tree based user interface allows end-users to quickly construct task plans.
Nguyen et al. describe ROS Commander as a user interface based on finite state machines for authoring task plans~\cite{nguyen2013ros}.
Similarly, Steinmetz and Weitshat~\cite{steinmetz2016skill} describe a graphical tool called RAFCON.

An alternate approach to direct task specification is to learn tasks from demonstrations. Wachter et al. automatically segment demonstrations to learn tasks~\cite{wachter2013action}. Alizadeh et al. learn skills which can be re-used according to a PDDL planner~\cite{ahmadzadeh2015learning}. Levine et al. proposed reinforcement learning methods for effectively learning individual skills with a demonstration as a prior~\cite{levine2015learning}.
In these cases, the end user still needs a way to connect individual skills.
Dianov et al. take a hybrid approach, using task graph learning to infer task structure from demonstrations and a detailed ontology~\cite{dianov2016extracting}.
Other recent work explored combining learned actions with sampling-based motion planning and a high-level task specification~\cite{paxton2016want}. 

\section{SYSTEM}\label{sec:system}

CoSTAR is a Behavior Tree-based user interface that aims to facilitate user interaction through a combination of an intuitive user interface, robust perception, and integrated planning and reasoning operations. It consists of Components, each of which is associated with a set of symbols, predicates, and operations. We generally expose operations to users as Behavior Tree leaf nodes, which can be parameterized and combined into a task plan capable of solving a wide range of problems. What follows is a brief high level overview of the CoSTAR system; for more detail see~\cite{paxton2017costar}. 

\begin{figure}[bt]
\centering
\includegraphics[width=0.9\columnwidth]{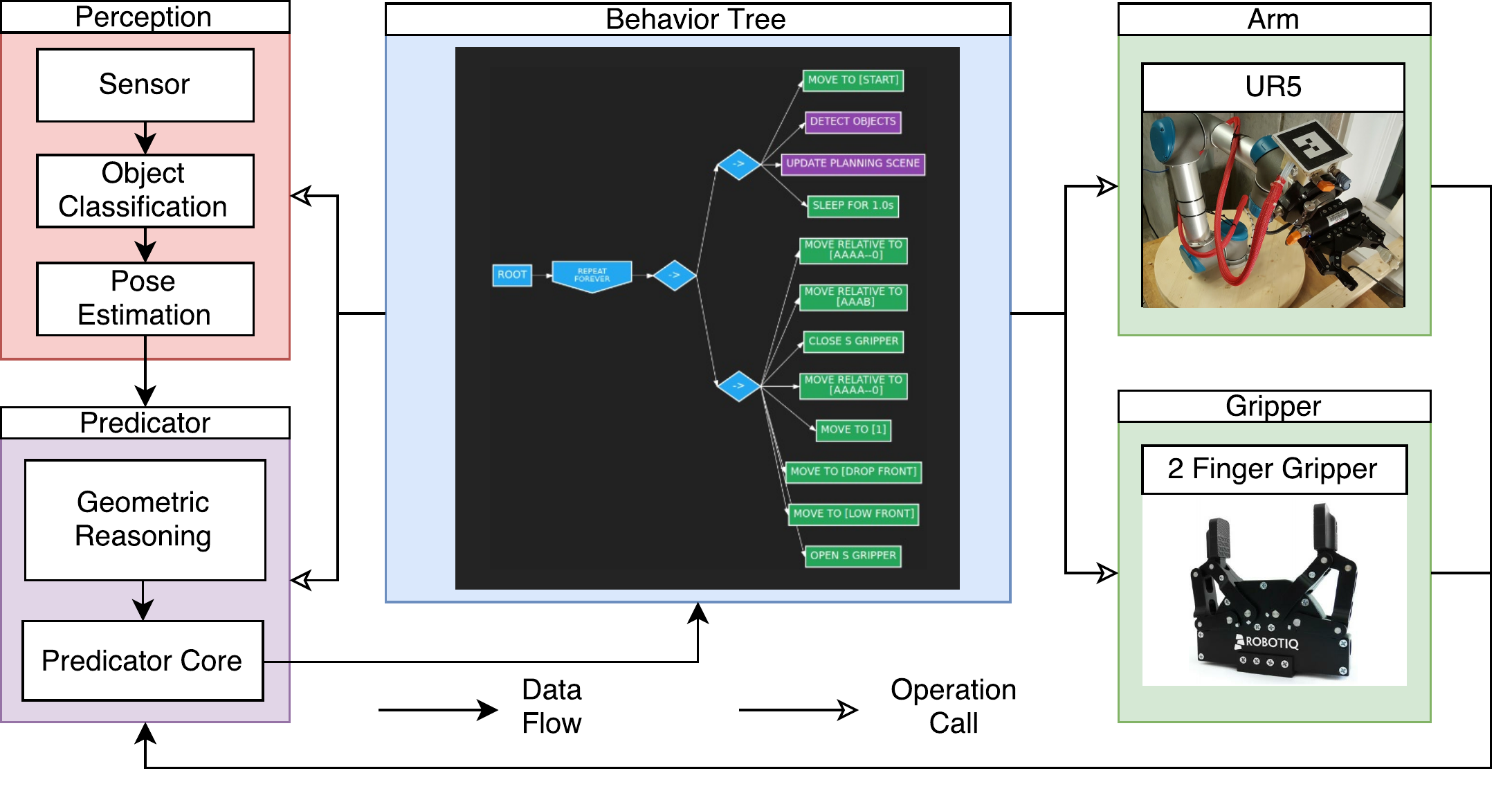}
\caption{A compressed overview of the CoSTAR system, based on the similar diagram in our previous work~\cite{paxton2017costar}.}
\label{fig:costar-system}
\end{figure}

\textbf{Components} are the composable elements of the CoSTAR system. A component is defined as a tuple $C = <I, O, p, s, u>$,
where $p = \{p_i\}_{i=1}^N$ is the set of predicates supported by $C$, $s = \{s_{i}\}_{i=1}^N$ is the set of symbols produced by $C$, and $u = \{u_{i}\}_{i=1}^N$ is the set of operations exposed by $C$. $I$ and $O$ are continuous input and output streams respectively.

\textbf{Symbols} $s$ are specific entities populated from continuous input data that the robot can act on, such as waypoints or candidate grasp positions.

\textbf{Predicates} $p$ are functions of a set of symbols $s_0,\dots,s_n$ and continuous input data $I$ that described some property of the world, i.e. $p(I,s_0,\dots,s_n) \rightarrow [\texttt{TRUE}, \texttt{FALSE}]$.
These describe relationships among objects, spatial occupancy, and object class membership.

\textbf{Operations} $u$ are specific actions that influence the world or update the robot's knowledge thereof. They can change the values associated with different symbols, and are generally exposed as Behavior Tree leaf nodes.

\begin{figure*}[t]
\centering
\includegraphics[width=1.9\columnwidth]{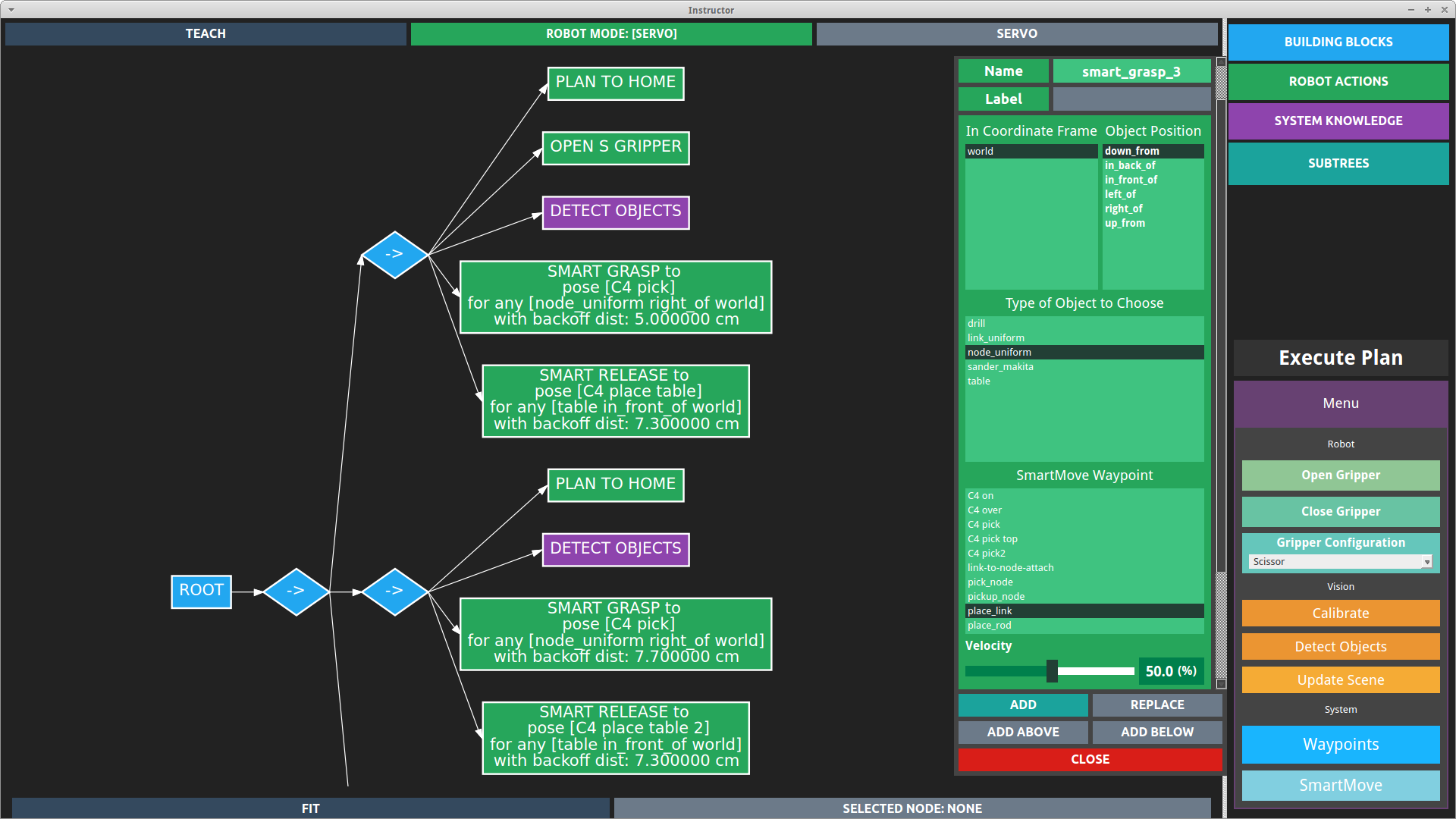}
\caption{The CoSTAR user interface displaying the Task 3 reference tree with \texttt{SmartGrasp} and \texttt{SmartRelease} actions.}
\label{fig:ui}
\end{figure*}

The most important components are the \texttt{Arm},  \texttt{Gripper}, \texttt{Perception}, and \texttt{Predicator} components, as shown in Fig.~\ref{fig:costar-system}, plus the \texttt{Instructor} user interface component shown in Fig.~\ref{fig:ui}. 
The primary extension explored in this paper is in the form of several new operations users can perform and algorithmically small but impactful improvements to existing components that improve accuracy and responsiveness.

\subsection{Predicator}

The Predicator component stores and tracks the robot's knowledge of the world. Predicator stores information on which objects have been detected, what types of objects they are, and how they relate to each other and the task at hand.
It is exposed to end users through the \texttt{KnowledgeTest} and \texttt{PoseQuery} operations.
The \texttt{KnowledgeTest} operation checks to see if a certain predicate is true. It is most often used for interactive tasks: it allows the user to determine if a particular region is occupied in order to create complex machine tending or collaborative behaviors.
The \texttt{PoseQuery} operation returns a list of goal frames that match a particular set of predicates, sorted by a cost function $c(\Delta q, \Delta t, \Delta R)$ over the distance between the current robot position and the goal position.

The default cost function is calculated as follows.
Let $W$ be a weight vector associated with joint angle difference $(\Delta q)$, translation$(\Delta t)$, and rotation $(\Delta R)$ respectively.
We add a cost term $\lambda$ associated with a projected collision with any objects in the world if our estimated inverse kinematics for the goal pose will be in collision.

\begin{equation}
\label{eqn:Cost}
\texttt{Cost}= W^T [\Delta q, \Delta t, \Delta R] +
\left\{
\begin{array}{lcl} 
0 & \text{no collision} , \\
\lambda & \text{collision}. 
\end{array} 
\right.
\end{equation}
In the current version of the CoSTAR UI, the predicate set input for the \texttt{poseQuery} operation is limited to geometric position and object class, so we can make queries like ``get grasp for any object left of the table pose where object is a node.'' See Alg.~\ref{alg:query} for the procedure used in \texttt{PoseQuery} operation.

\begin{algorithm}[bt!]
\caption{Pose Query: generates a list of possible goals given a set of predicates and detected objects based on estimated inverse kinematics at goal.}
\label{alg:query}
\begin{algorithmic} 
\Function{PoseQuery}{Detected objects $\mathcal{O}$,
     predicates $\mathcal{P}$, current joint position ${q}$, planning scene world $\mathcal{W}$,
    relative end effector pose ${T}_{EE}$, cost function $c$}
     \State $\mathcal{G} = \emptyset$ \Comment Empty set of possible goal poses
     \State $T_{c} = $ \Call {SolveForwardKinematics}{$q$}
     \For {$o, T_o \textbf{ in } \mathcal{O}$}
        \If {$p(o)$} \Comment Object pose matches the predicate
         \For {$T_s \textbf{ in }$ \Call {GetPoseSymmetries}{$o$}}
            \State $T_{sym} = T_o \cdot T_s \cdot T_{EE}$
            \State $valid = $ \Call {CollisionsCheck}{$T_{sym}, \mathcal{W}, q$}
            \State $q_i = $ \Call {SolveIK}{$T_{sym}, q$}
            \State $c_{q,t,R} = c(q_i,q, T_{sym}, T_{c}, valid)$
            \State \Call{Insert} {$\mathcal{G}, \{ o , T_{sym} , c_{q,T,R}$}
         \EndFor
        \EndIf
     \EndFor
     \State \Return \Call {Sort}{$\mathcal{G},c_{q,t,R}$}
\EndFunction
\end{algorithmic}
\end{algorithm}

\subsection{Arm}

The \texttt{Arm} component handles motion planning and execution, and ties in closely with the \texttt{Predicator} component to expose more advanced operations.

Each arm is associated with a unique \texttt{home} symbol. This symbol represents a position where the camera can perceive the workspace unobstructed.
We add the \texttt{PlanToHome} and \texttt{MoveToHome} operations separate from the \texttt{Move} and \texttt{PlannedMove} operations in order to explicitly expose this functionality to end users, and make it more clear to them when the robot will be able to cleanly perceive its environment.


We added \texttt{SmartGrasp} and \texttt{SmartRelease} actions to the existing system as well. These can be thought of as small sequences of commands that query the robot's knowledge of the world for objects matching some set of conditions using \texttt{PoseQuery}. The \texttt{PoseQuery} function will generate a sorted list of possible goals for the object of interest.
The \texttt{Arm} uses the resulting sorted list to generate motion plans in order of preference via the RRT-Connect algorithm.

\texttt{SmartGrasp} and \texttt{SmartRelease} have one additional parameter: a backoff distance. This can be set between 1 and 10 cm. When planning the move, the arm will first move to a position that is this backoff distance away from the trained pose before moving directly in to the final position. For \texttt{SmartGrasp} this backoff distance is computed in the gripper x frame, and for \texttt{SmartRelease} it is computed in the world z axis so objects can be stacked or placed on the table.
These new operations combine several capabilities from different components of the CoSTAR system. To train these operations users first select \texttt{DetectObjects} on \ref{fig:ui} then select an object that they wish to manipulate from a list and name the associated pose.

\subsection{User Interface}\label{sec:ui}

The user interface of the CoSTAR system includes a Behavior Tree-based task editor and a 3D visualization of the robot's workspace.
The editor shown in Fig.~\ref{fig:ui} allows the end user to combine and parameterize operations exposed to the user by all of these different components. The menu in the lower right lists useful training operations including \texttt{DetectObjects}, \texttt{OpenGripper}, and \texttt{CloseGripper}. Object information is updated via the \texttt{DetectObjects} primitive operation, and frame based knowledge is created via \texttt{Waypoints} and \texttt{SmartMove}.

\begin{figure*}[bt]
    \centering
    \begin{subfigure}[b]{\columnwidth}
        \includegraphics[width=\columnwidth]{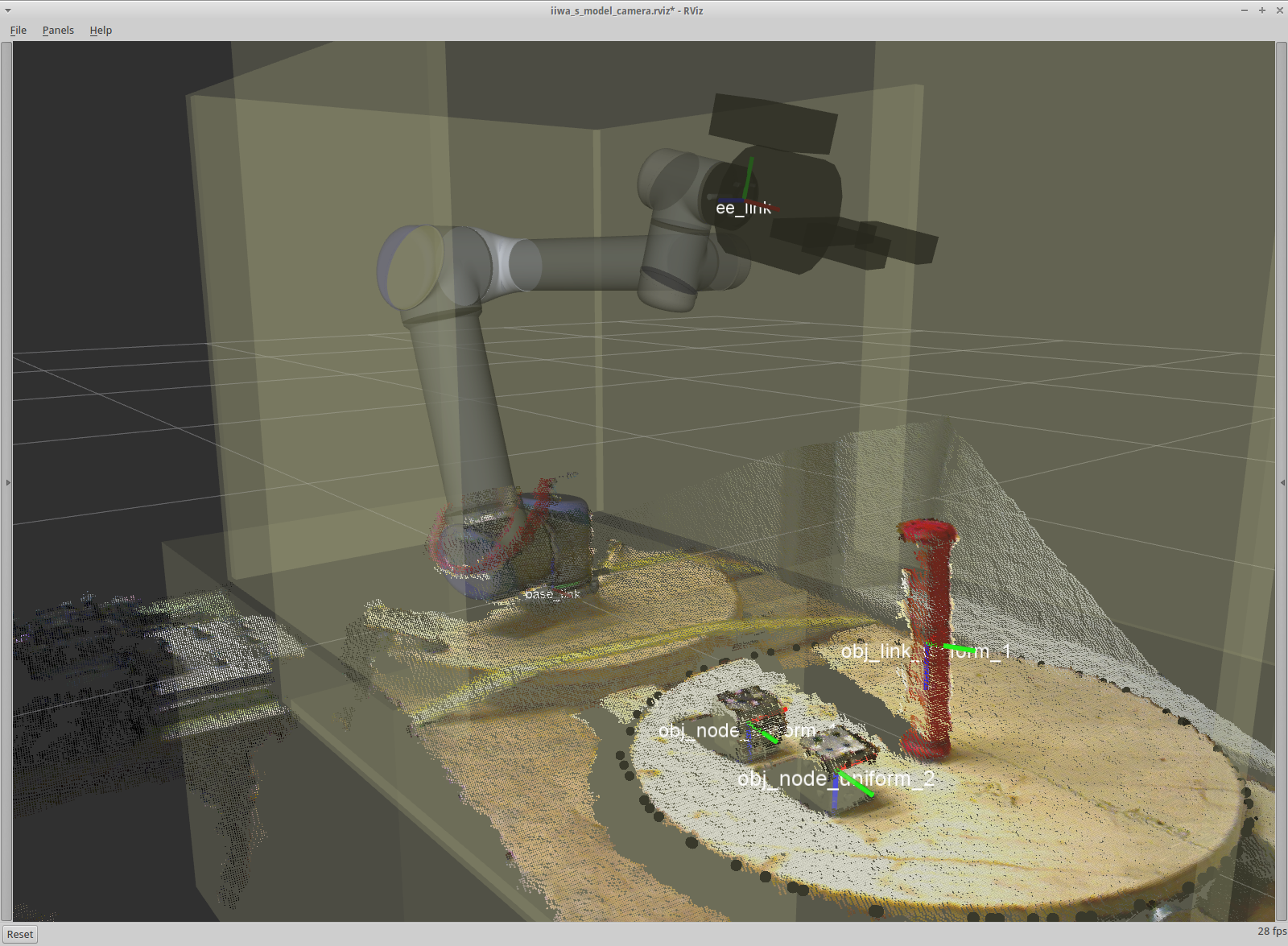}
        \caption{}
        \label{fig:rviz-before}
    \end{subfigure}
    \begin{subfigure}[b]{\columnwidth}
        \includegraphics[width=\columnwidth]{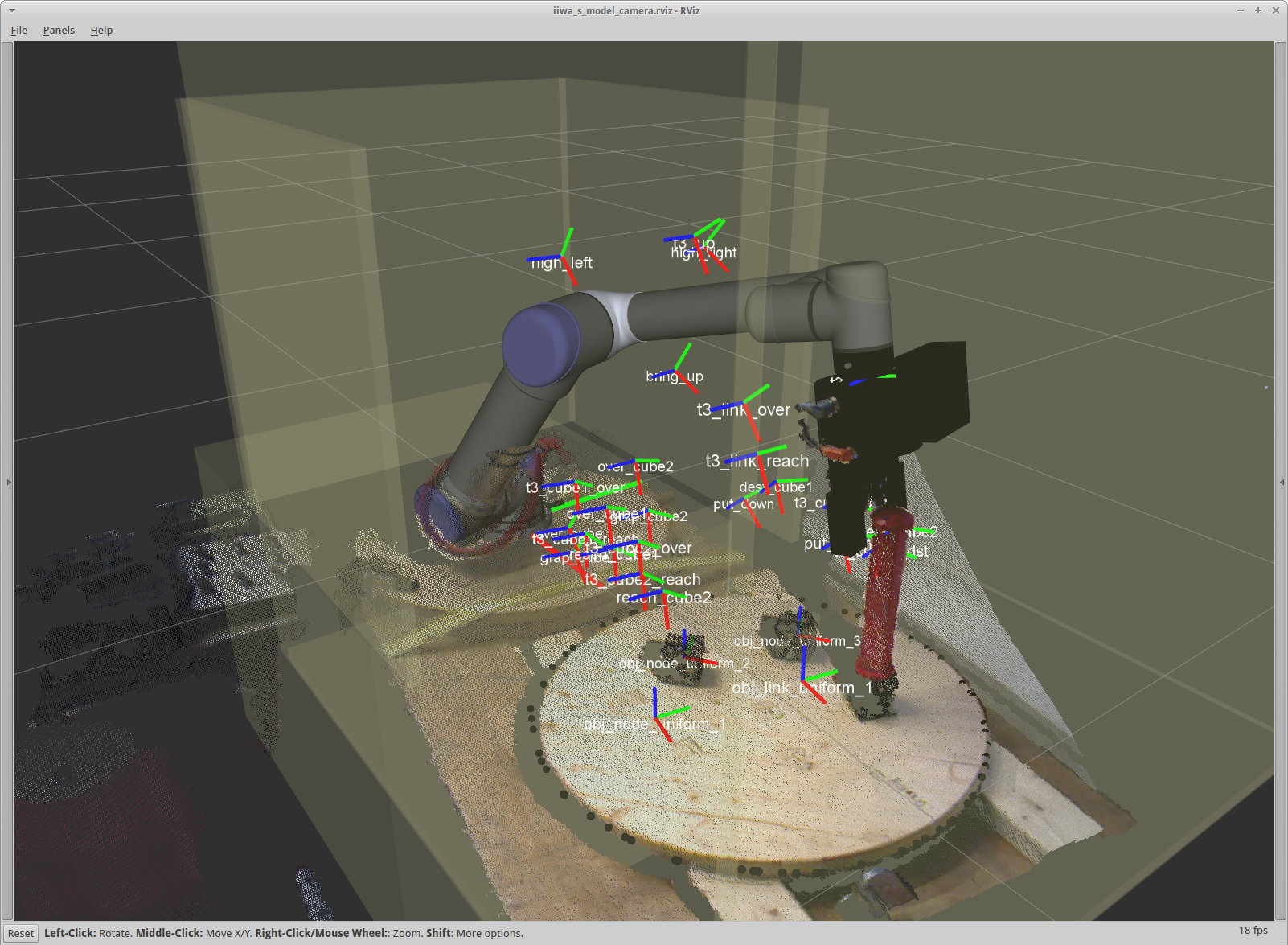}
        \caption{}
        \label{fig:rviz-after}
    \end{subfigure}
    \caption{Displaying what the robot knows. \ref{fig:rviz-before} shows collision boundaries around the world and detected objects; \ref{fig:rviz-after} shows the large number of waypoints taught by a particular user.}
    \label{fig:rviz}
\end{figure*}

The ROS~\cite{quigley2009ros} RVIZ interface is the second screen of the user interface shown in Fig.~\ref{fig:rviz}. This displays what the robot knows including object positions and waypoints, demonstrated coordinate frames, planned trajectories, collisions, workspace boundaries, and the planning scene which represents detected objects. When the robot has picked up an object we also update and display attached objects, as is done in the \texttt{SmartGrasp} and \texttt{SmartRelease} operations.

\section{USER STUDY}

We conducted a preliminary user study of the CoSTAR system to assess the effectiveness of the CoSTAR system in supporting users to construct complex task plans and the user experience resulting from the current interface.
Our goals were to gain a better understanding of (1) the effectiveness of the CoSTAR system in supporting users to construct complex plans, (2) the user experience resulting from the current interface, and (3) the relative utility of different CoSTAR operations to novice end-users.

\subsection{Study Design}\label{studydesign}
The study involved four study conditions that represented the set of abilities of the CoSTAR system discussed earlier and shown in Table~\ref{table:conditions}: (1) baseline, (2) planning, (3) perception, and (4) SmartMove. Participants were randomly assigned to one of these conditions and asked to complete the study tasks using the set of abilities afforded by the condition.

\textbf{Condition 1} is meant to represent a baseline system similar to that of the Universal Robot UR5 or the Rethink Robotics Sawyer. In this system, users cap program grasps and waypoints, but do not have access to perception or motion planning. 

\textbf{Condition 2} adds planning: the user can detect all objects in the scene and the robot will plan trajectories that avoid collisions and joint limits. They must use the \texttt{DisableCollisions} operation with the ID of any object they wish to manipulate to remove it from the motion planning scene. 

\textbf{Condition 3} tests the utility of simple perception: CoSTAR can detect objects and assign them IDs, and users  can track waypoints relative to detected object positions.

\textbf{Condition 4} includes access to the \texttt{SmartGrasp} and \texttt{SmartRelease} operations discussed in Sec.~\ref{sec:system}. These operations perform a query to select possible objects based on semantic information such as relative position and object class. In all cases, poses are taught via manipulating the UR5 robot while it is in freedrive mode, and adding an appropriate pose to CoSTAR's memory via a UI.


\begin{table}[bt]
\centering
\caption{The four conditions correspond to four sets of actions, each highlighting different characteristics of the user interface.}
\begin{tabular}{ c l  c  c }
\hline
Condition & Description & Actions & Knowledge \\
\hline
1 & Baseline & Move to Home & \\
& & Move to Waypoint & \\
\hline 
2 & Planning & Plan to Home & Detect Objects \\
& & Plan to Waypoint & Enable Collisions \\
\hline
3 & Perception & Move to Home & Detect Objects \\
& & Move Relative to Object & \\
\hline
4 & SmartMove & Plan to Home & Detect Objects \\
& & Smart Grasp & \\
& & Smart Release & \\
\hline
\end{tabular}
\label{table:conditions}
\end{table}

\begin{figure*}[hbt]
    \centering
    \begin{subfigure}[b]{\columnwidth}
        \includegraphics[width=\columnwidth]{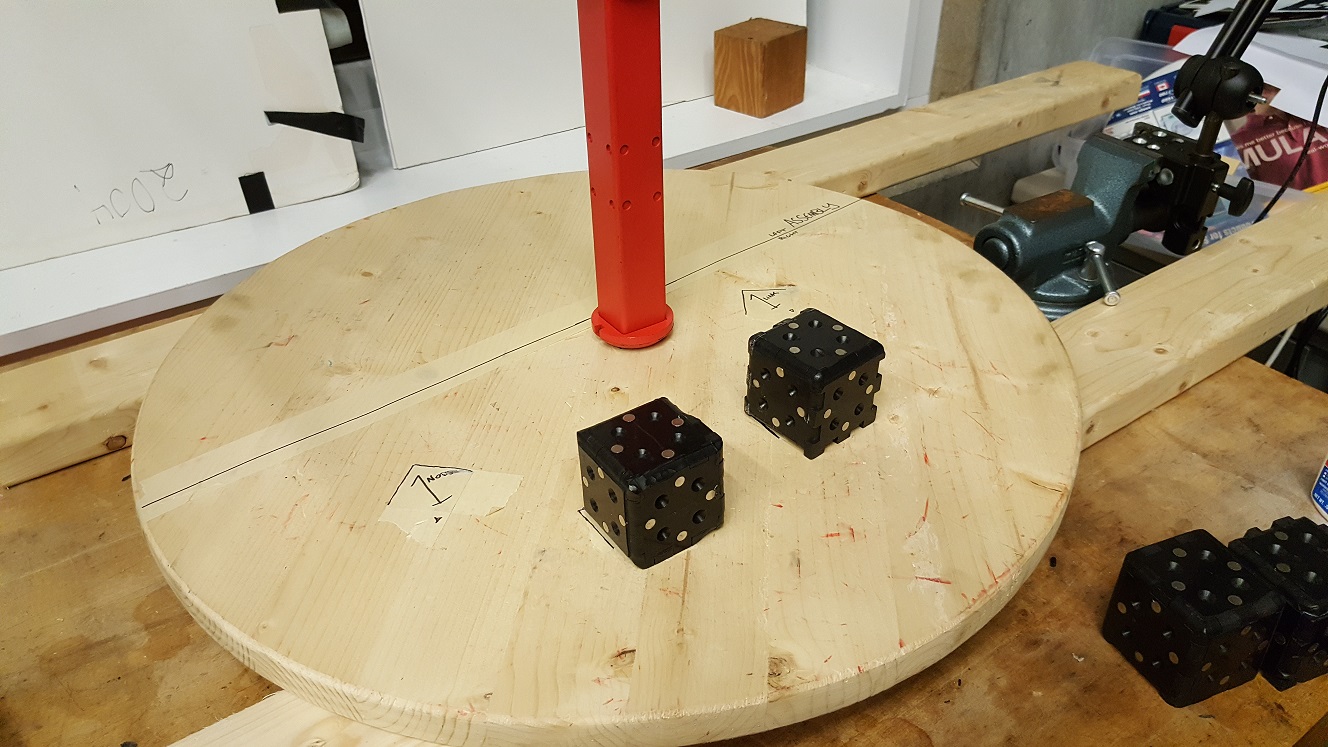}
        \caption{Task 2: move two nodes with obstacle}
        \label{fig:task2}
    \end{subfigure}
    \begin{subfigure}[b]{\columnwidth}
        \includegraphics[width=\columnwidth]{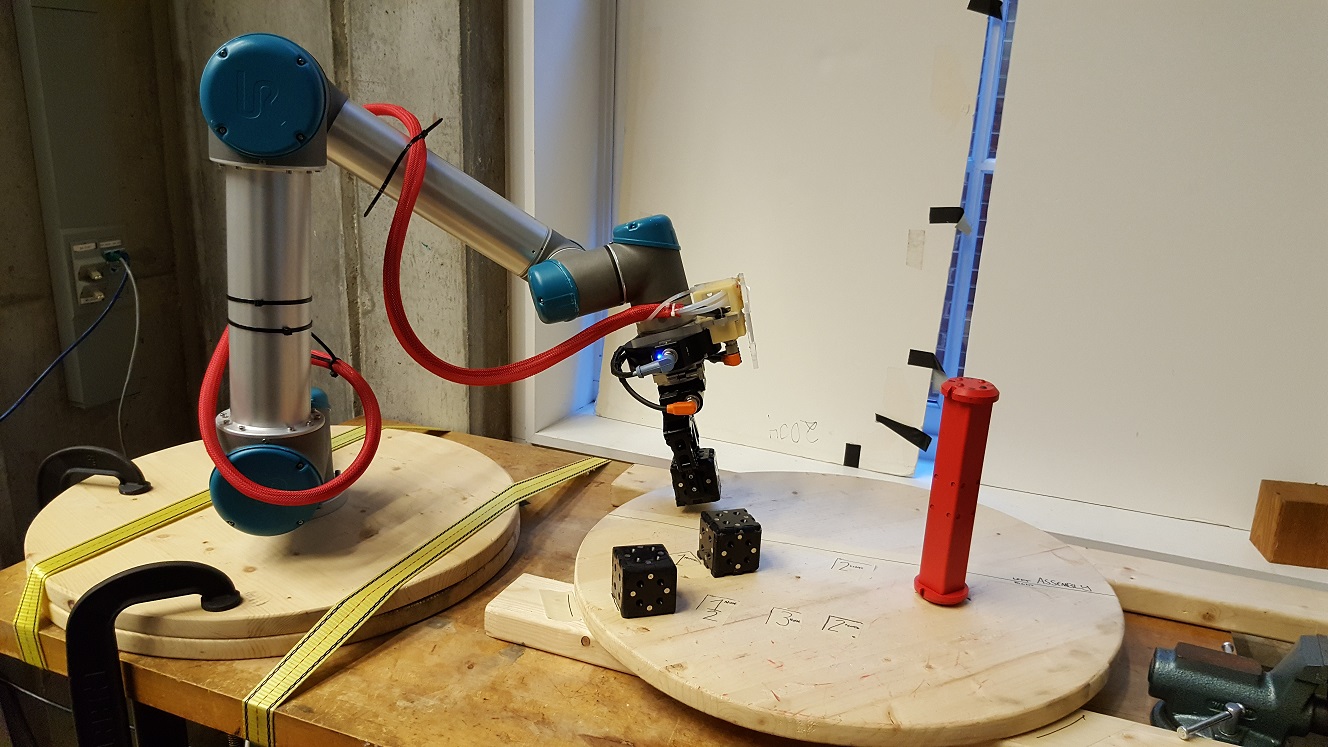}
        \caption{Task 3: move any two nodes and stack link}
        \label{fig:task3}
    \end{subfigure}
    \caption{Users were presented with three different configurations of a set of blocks, two of which are shown here.}
    \label{fig:tasks}
\end{figure*}

We presented participants with three ``pick-and-place'' tasks with increasing complexity, two of which are shown in Fig.~\ref{fig:tasks}. The tasks were designed to enable participants to incrementally learn and put into practice how each UI component works, how the robot responded to user commands, and how to build task plans using specific technologies such as perception and planning.
Performing all three tasks required participants to move square blocks in different configurations from the right to the left of a circular workspace without knocking over an obstacle: a red ``link.'' The last task additionally required them to pick up the red link and connect it to one of the blocks, representing an additional move action that added further complexity to the pick-and-place task.

\textbf{Task 1} asked participants to move two blocks from the right side of the workspace to the left. The goal of this task was to determine whether participants could apply the knowledge from expert demonstration to teach the robot themselves, with no notable challenges other than repeating the instructions for a new node. The obstacle was introduced to the world, but it was far enough away from the blocks that participants did not need to actively avoid the obstacle. They were given the single node solution created by the experimenter during demonstration as a starting point, which would allow them to move at least one block even if they did not modify the task plan.

\textbf{Task 2} required participants to similarly move two blocks from the right to the left, although one of the blocks was placed in a different position from the previous task and the obstacle was placed closer to the two objects (Fig.~\ref{fig:task2}). As in Task 1, Participants were given the solution to the previous task. Given that one of the nodes was in the same position as in Task 1 and in the original demonstration, participants could run the tree they were given and accomplish half of the task.

\textbf{Task 3} presented participants with three blocks all of which were in positions different from previous tasks. The link was moved farther away again, and participants were asked to move two blocks of their choice and to pick up the link and place it on top of one of these blocks. This configuration is shown in Fig.~\ref{fig:task3}.


\subsection{Study Procedure}
The user study involved three phases.
In the first phase, which took 10-15 minutes, an experimenter demonstrated the use of the system and answered any questions that the participant might have.
During the demonstration, the experimenter constructed an example task that involved moving one block from the right side of the workspace to the left.

In the second phase of the study, participants were provided with 15 minutes to perform the three tasks. The last phase of the study involved participants filling out a questionnaire that included the System Usability Scale \cite{albert2009beyond} and answering a set of interview questions. This procedure was approved by the Johns Hopkins University Institutional Review Board (IRB) under protocol \#HIRB00005268.

\subsection{Participants}

The preliminary study included five participants, two assigned to Condition 1 and one for each of the other conditions. The participants were undergraduate and graduate students with experience in robotics, engineering, and computer science, who represented our system's target group of non-expert but technically savvy end-users, such as manufacturing engineers, laboratory technicians, and so on..

\begin{table*}[thbp]
\centering
\caption{Results from each of the four conditions on the three tasks. User 1B did not complete Task 1 or 2 in time.}
\begin{tabular}{| c  l c | c | c | c | c | c | c |}
\hline
Condition & Description & User & Task 1 Time & Task 2 Time & \# Task 3 Parts Moved & SUS Score \\
\hline
\hline
1 & Baseline & A & 7:38 & 8:12 & 1 & 75 \\
1 & Baseline & B & N/A & N/A & 2 & 82.5 \\
2 & Planning & C & 13:08 & 9:22 & 3 & 75 \\
3 & Perception & D & 14:46 & 15:00 & 3 & 57.5 \\
4 & SmartMove & E & 11:00 & 3:08 & 3 & 67.5 \\
\hline
\end{tabular}
\label{table:task-results}
\end{table*}

\section{Results}

\begin{figure}[bt]
\centering
\includegraphics[width=\columnwidth]{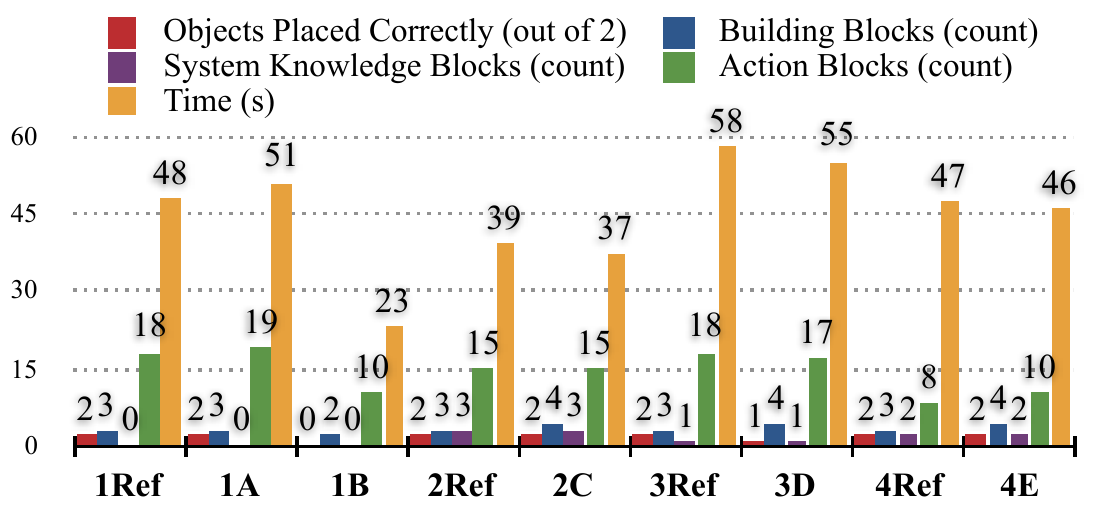}
\caption{Task 1, ``move two nodes'': performance comparison of solution trees  against a reference implementation. User labels indicate each test condition number and unique user. To succeed there must have been two objects placed.}
\label{fig:saved-tree-comparison}
\end{figure}

\begin{figure}[bt]
\centering
\includegraphics[width=\columnwidth]{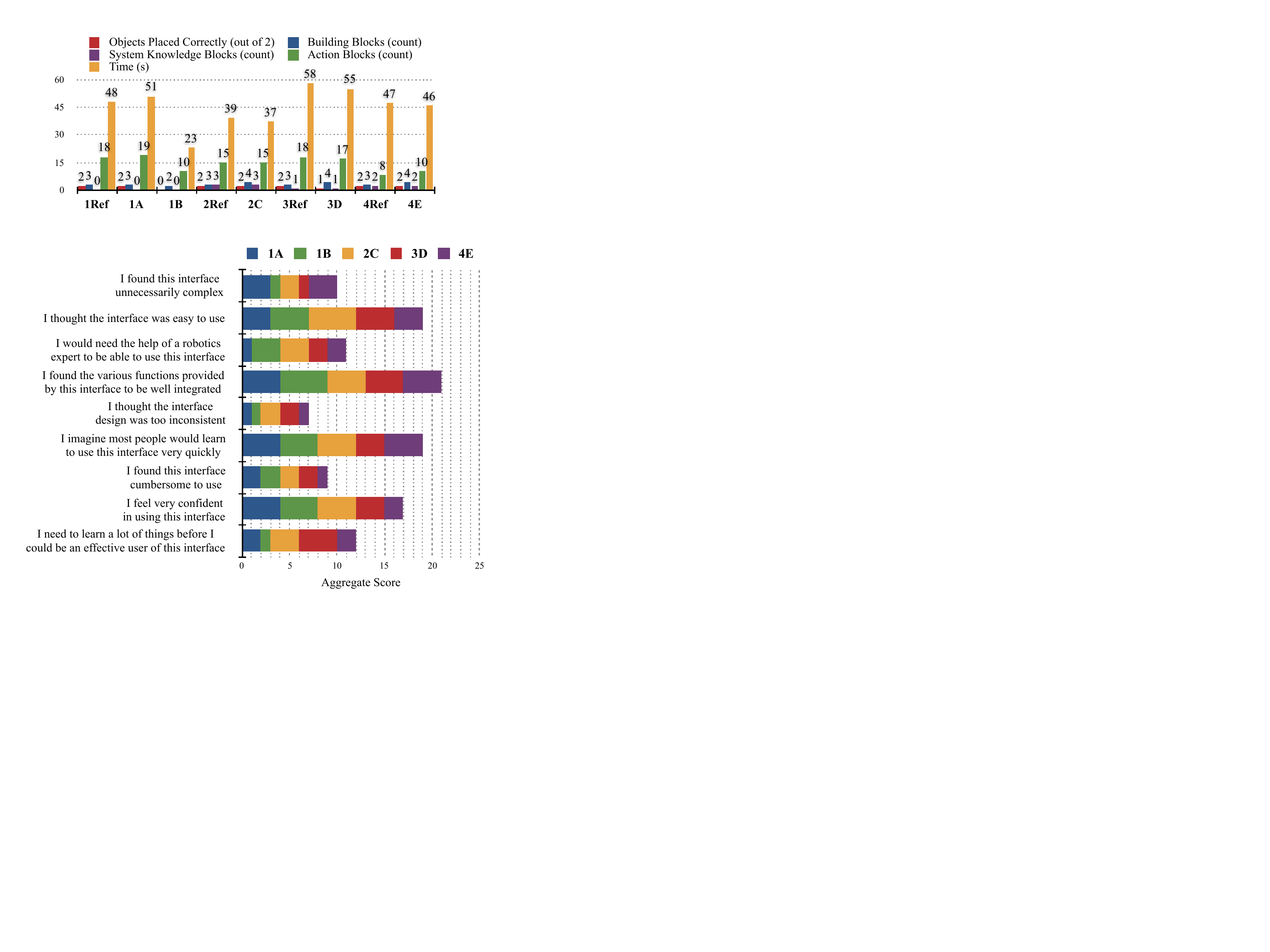}
\caption{Subjective user assessment of perceived system usability.
Scores from each user are from 1 ``strongly disagree'' to 5 ``strongly agree.''}
\label{fig:sus}
\end{figure}

Figures~\ref{fig:saved-tree-comparison} and \ref{fig:sus} and Table~\ref{table:task-results} summarize all the task performance and usability data collected in the study. Below, we discuss the main findings from our analysis of this data.

\subsection{Task Performance} Table II shows time to completion for each of the four users who completed Tasks 1 and 2. None of the participants were able to complete Task 3 successfully. Interestingly, the participant in Condition 1 was able to understand the system and use it the fastest; however, this participant was unable to leverage this understanding to more quickly complete the next task. In contrast, the participant in Condition 2 took longer to complete the first task but showed improvement on the second, where the addition of motion planning meant that they needed to rethink their approach to a lesser degree. 

The participant in Condition 4 was able to quickly adapt to a new scenario; they merely taught a new grasp for the second node. Note that this re-teaching step was not necessary and was a result of a misunderstanding by the participant.

\subsection{Failures} The two most common failures in our trials occurred when concepts related to perception or to multiple coordinate frames were unclear. First, the system makes no assumptions about initial object configurations and can assign different IDs for a single object across different detections. Thus, assuming a fixed relationship between position and ID results in unpredictable behavior. Second, the motion planner will return a failure if a goal is too close to the table or to an object unless \texttt{SmartMove} is used or collisions have been disabled with the \texttt{DisableCollisions} operation. Additional points of confusion included the misunderstanding that \texttt{SmartGrasp} operations had to be re-taught for every object of the node class and that \texttt{SmartRelease} operations did not need to be re-taught for every new node position on the workspace.

Participants in the two conditions that involved perception (Conditions 3 and 4) were unclear on when knowledge they provided to the robot would generalize and when it would not. On the other hand, participants in the conditions that did not involve perception (Conditions 1 and 2) found the robot to be easy to manage and predictable but expressed frustration by the degree to which they had to re-teach it.

\subsection{System Use} Figure 5 shows system use by the trees constructed by the demonstrator and study participants. We note that, of the successful trials, the solutions from Condition 2 were concise trees that used motion planning to generate efficient trajectories to goals. Condition 4 used fewer action blocks but took more time because of the need for perception. However, the solution from Condition 4 generalizes to many different environments and configurations of the blocks. This generalization is also supported by the performance increase between Tasks 1 and 2 in Condition 4 shown in Table~\ref{table:task-results}.

\subsection{Perceived Usability} Data from the System Usability Scale (SUS)~\cite{albert2009beyond} indicate that, in general, users found the baseline system to be highly usable. However, the added capabilities reduced the ease of use to average or low usability levels.
Other users found the integration with the perception system to be confusing, and were unclear how to answer the question ``I think I would use this interface frequently.'' 

\subsection{Qualitative Observations and Interview Data} The qualitative data collected through interviews indicate that users found the behavior tree, sometimes referred to as the "diagram" by users, to be particularly straightforward and easy to understand. Participant comments indicate that nodes changing colors as they execute or are completed might be contributing to this clarity. One participant expressed that the tree they constructed did exactly what they wanted it to do. 

\textbf{Condition 1.} We found that participants in Condition 1 understood all of the core concepts and were able to complete their tasks very quickly, although we observed that one of the participants accidentally deleted most of their tree before it was evaluated and thus did not complete Task 1. Based on this observation, an important improvement will be to provide a mechanism for users to recover from such errors.

\textbf{Condition 2.} The participant in Condition 2, where the perception capability was disabled, suggested that perception would be a valuable addition to the system and that capabilities of collapsing, saving, and reloading subtrees would be valuable additions. Condition 2 also made the transition from Task 1 to Task 2 simpler when compared to the Condition 1, because the participant in Condition 2 did not need to teach a set of intermediate poses to avoid colliding with the obstacle. However, while there was some improvement between the tasks, the participant in this condition still found the task to be relatively difficult.

\textbf{Condition 3.} The third condition involved perception but not planning, and the participant in this condition was asked to teach the robot positions relative to coordinate frames generated by the motion planner. Performing the task this way required an understanding of what motions needed to be taught in which frame of reference and how to do this, which was not immediately obvious to the participant.

\textbf{Condition 4.} The participant in Condition 4 appeared to be able to generalize the task plan to a new context much more easily than others did. Despite the improvement in their task performance, this participant was not conﬁdent in the resulting solution and found the procedure of adding a new SmartMove unnecessarily complex and confusing. SmartMoves take the form of a query: "for any object and associated pose such that $p_1, p_2, \dots , p_N$ are true, move to pose." We speculate that of the participant's confusion was due to the names of the fields in the SmartMove menu and the process of filling out the conditions in the query. We asked participants to make two different conditions: a geometric spatial condition ($p_1$ = "find any object on the right side of the world") and an object class condition ($p_2$ = object is a node"). To make the user interface easier to follow, we renamed several elements of the SmartMove UI: "Reference" became "in coordinate frame;" "select region” became “object position,” and “select object type” became “type of object to choose.” These changes are reflected in Fig. \ref{fig:ui}.

Users in this pilot study found some aspects of the experimental design confusing or frustrating. In particular, they disliked receiving new trees after every task, as they were unsure what they could re-use or replace. This observation reveals two requirements for the current user interface. First, users should be able to replay only a specific part of the tree. Second, users need a clear visualization of what they have taught the robot.

We also observed that participants found the large number of waypoints provided before the task to be confusing, as they could not tell which ones would be useful for a given task. This confusion only aggravated the problems many users had with re-using trees they did not program themselves. In fact, several users chose to delete existing trees entirely and start from scratch, which we did not intend.

Finally, the interview data indicated several user requests for additional functionality or usability improvements, including the ability to program nodes with code (an existing capability that was outside the scope of the study), better clarity on what frames really meant and how positions relative to them were deﬁned, and more extensive use of the RGBD sensor, which helped participants even in conditions that did not involve perception, as it visualized coordinate frames and helped users understand mappings to the real world.


\section{DISCUSSION}
\begin{figure*}[t]
    \centering
    \begin{subfigure}[b]{0.5\columnwidth}
        \includegraphics[width=\columnwidth]{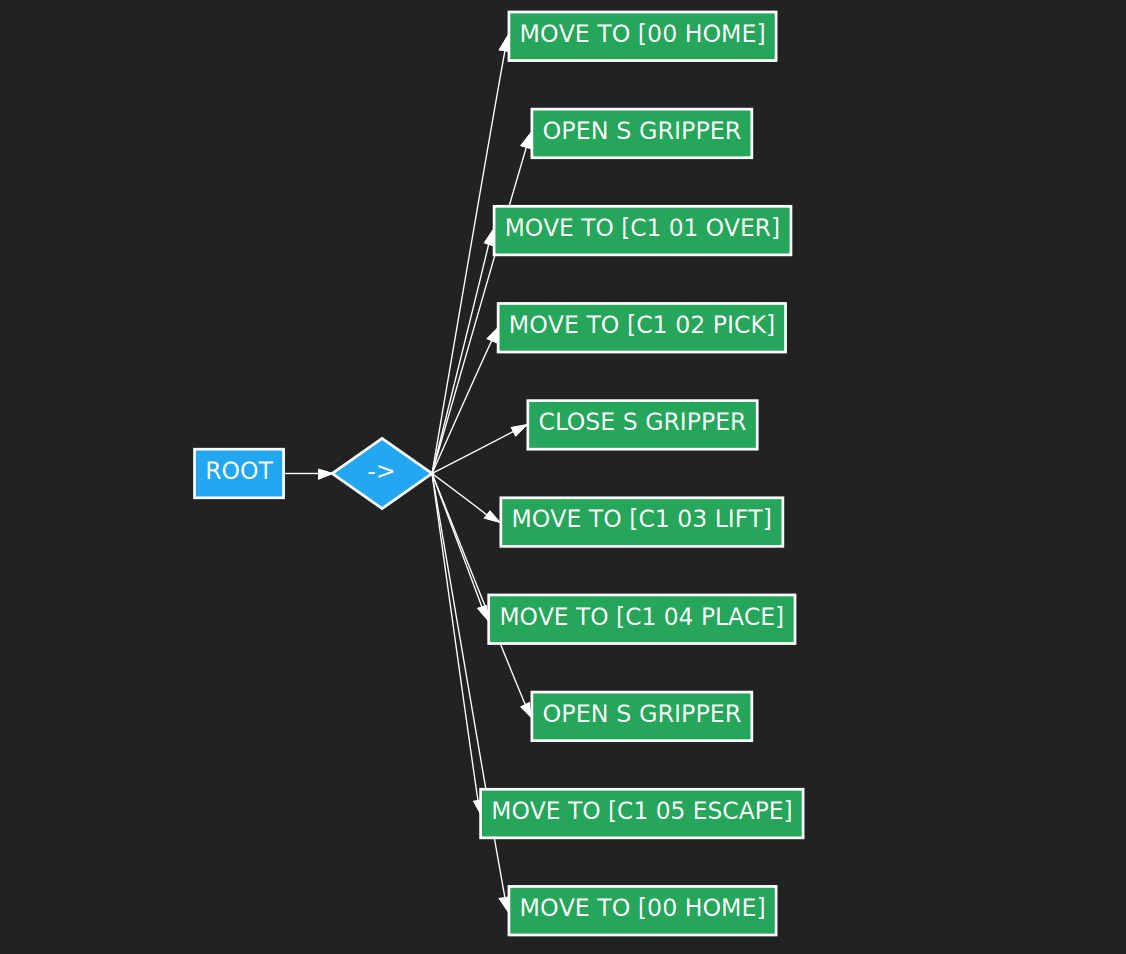}
        \caption{Condition 1: Baseline}
        \label{fig:Condition1}
    \end{subfigure}
    \begin{subfigure}[b]{0.5\columnwidth}
        \includegraphics[width=\columnwidth]{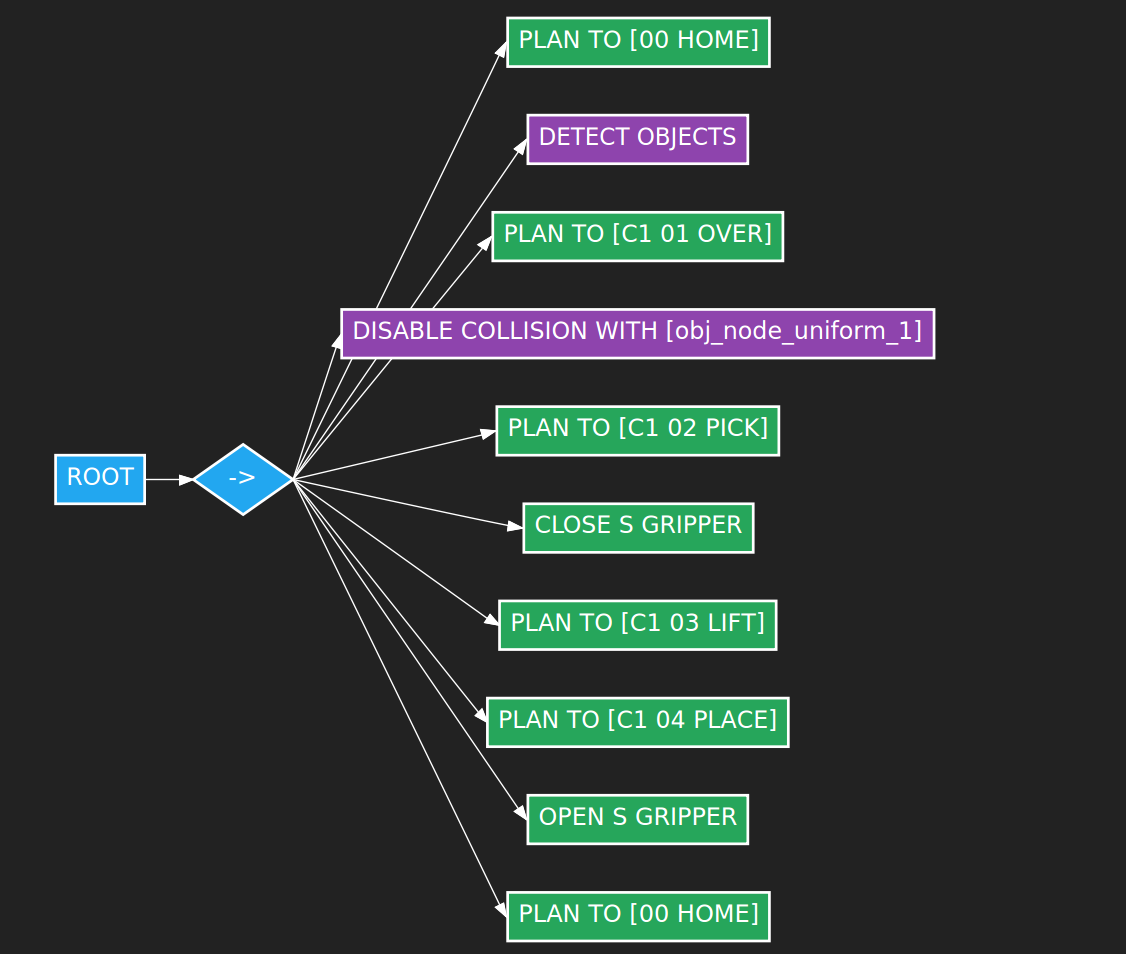}
        \caption{Condition 2: Planning}
        \label{fig:Condition2}
    \end{subfigure}
    \begin{subfigure}[b]{0.5\columnwidth}
        \includegraphics[width=\columnwidth]{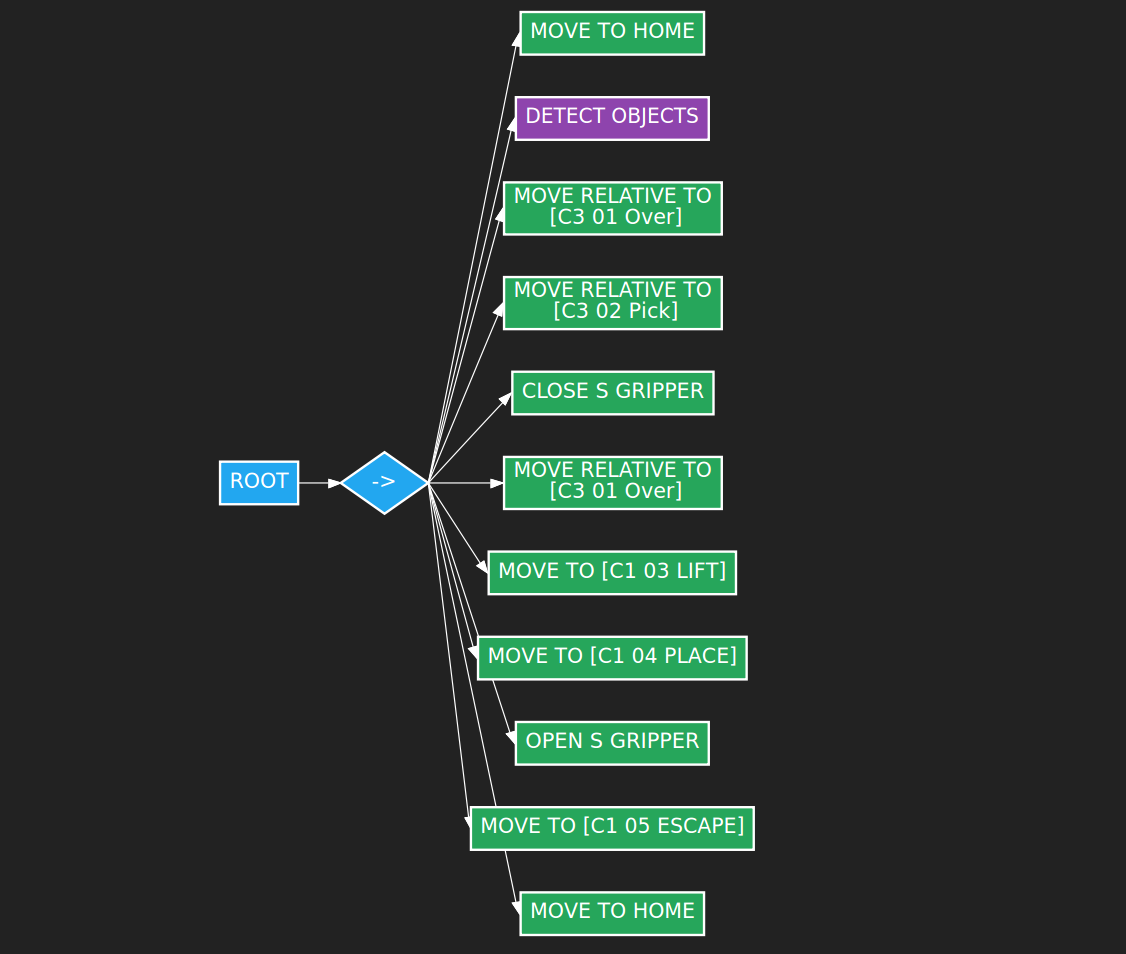}
        \caption{Condition 3: Perception}
        \label{fig:Condition3}
    \end{subfigure}
    \begin{subfigure}[b]{0.5\columnwidth}
        \includegraphics[width=\columnwidth]{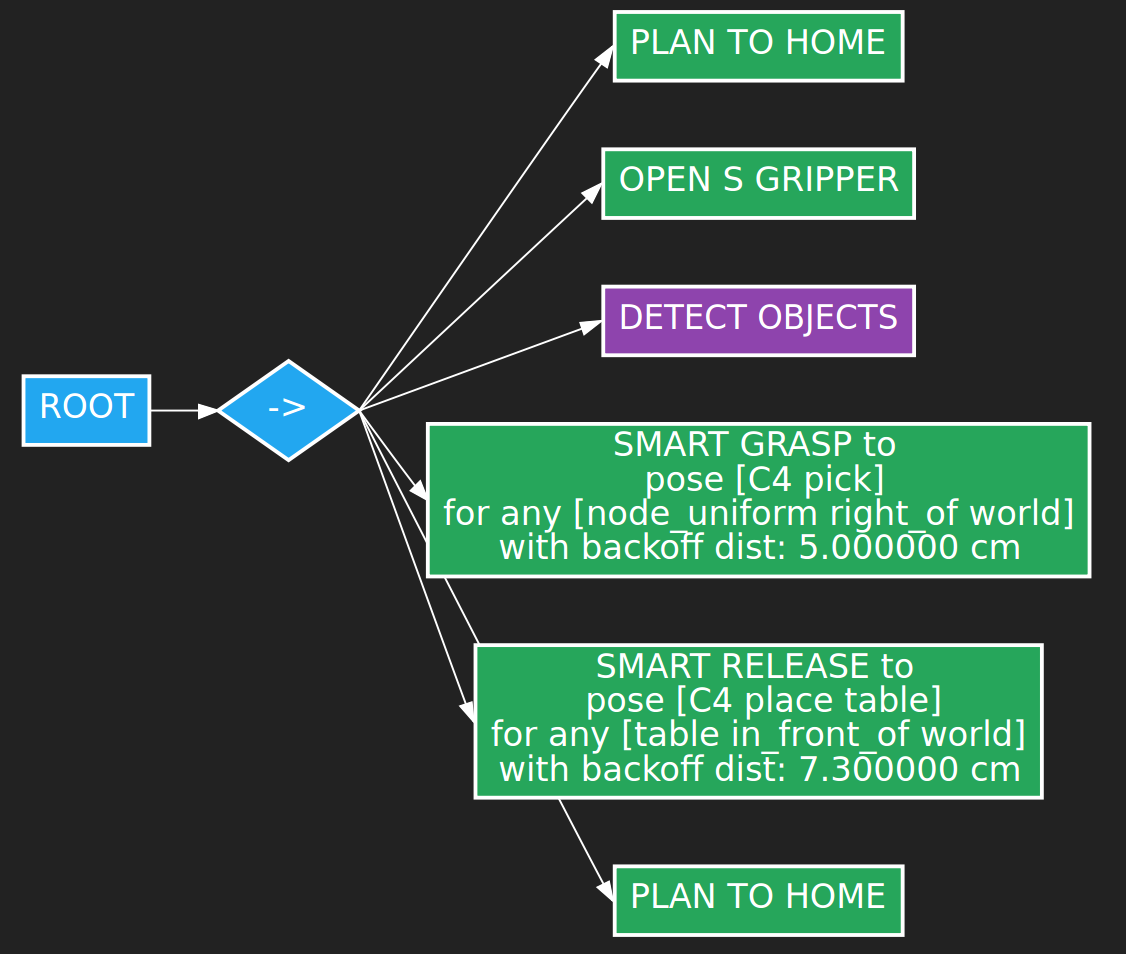}
        \caption{Condition 4: SmartMove}
        \label{fig:Condition4}
    \end{subfigure}
    \caption{The CoSTAR operations were enabled and disabled under four different Conditions to test the value of each capability across users as described in \ref{studydesign}.}
    \label{fig:Conditions}
\end{figure*}

There are three lessons from this preliminary study. First, these results show that novice users find Behavior Trees to be a practical and effective means of defining a robot program. The findings also show that integrating planning, perception, and simple reasoning makes programming robots faster, more effective, and more general.
Finally, they reveal the importance of helping users build more accurate mental models of robot capabilities.
In particular, we see that perception is not useful to our participants if they cannot communicate effectively with the robot.
The main disconnect between novice users and robotic systems is in understanding what the robot will do and why it will do it.
We argue that such mental-model problems are the source of the relatively poor performance in Condition 3 compared to 2 or 4. Condition 3 is perhaps the least intuitive set of conditions (though not the most complex to specify). The SmartMove operations in Condition 4 were the most complex to instantiate, but resulted in reliable behavior that generalized well to different conditions, and result in concise, readable trees (see Fig.~\ref{fig:Condition4}).

In the future, we plan to provide users with support for building off of the same tree when completing successively more complex tasks. We also plan to refine the user inteface, in particular to make it clearer what different elements of the SmartMove user interface mean.

\section{CONCLUSIONS}



We described a preliminary between-participants study exploring the effectiveness and usability of the CoSTAR system for instruction of collaborative robots.
Users were able to effectively program the robot to solve object manipulation tasks with Behavior Trees under various study conditions that tested particular aspects of the system.
Adding in more advanced capabilities allowed users to solve more challenging problems and to build more reliable solutions, but at a notable cost to added complexity.
Finally, we observed that most issues arise from a disconnect between what the user thinks an operation means and what the robot will actually try to do, or from cases where an interface was confusing to the user.
In our future work we plan to address concerns raised with the study by allowing users to keep and build off of their previous plans, and providing hands-on instruction for the first 10-15 minutes rather than providing a demonstration.
We also aim to make CoSTAR a more reliable system, expanding the suite of reasoning and planning capabilities.


\bibliographystyle{IEEEtran}
\bibliography{kuka,lfd,planners,software,vision,costar,psych}

\end{document}